\pdfoutput=1

\documentclass[11pt]{article}

\usepackage[final]{acl}

\usepackage{times}
\usepackage{latexsym}

\usepackage[T1]{fontenc}

\usepackage[utf8]{inputenc}

\usepackage{microtype}

\usepackage{inconsolata}
\usepackage{enumitem}
\usepackage{times}
\usepackage{latexsym}
\usepackage{amssymb}
\usepackage{amsmath} 
\usepackage{amstext}
\usepackage{booktabs}
\usepackage{enumerate}
\usepackage{graphicx}
\usepackage{subfigure}
\usepackage{xspace}
\usepackage{float}
\usepackage{bbm}
\usepackage{bm}
\usepackage{multirow}
\usepackage{booktabs}
\usepackage{color}
\usepackage{framed}
\usepackage{stfloats}
\usepackage{iitem}
\usepackage{threeparttable,booktabs}
\usepackage{pifont} 

\usepackage{color, colortbl}
\definecolor{Gray}{gray}{0.9}
\definecolor{LightCyan}{rgb}{0.88,1,1}

\definecolor{bluencs}{rgb}{0.0, 0.53, 0.74}
\newcommand{\cmark}{\ding{51}}%
\newcommand{\xmark}{\ding{55}}%

\newcommand{\modelname}[1]{\textsc{Adam}}

\title{\textsc{Adam}: Dense Retrieval Distillation with Adaptive Dark Examples}
\author{Chongyang Tao$^{1}$\thanks{\ \ Equal Contribution.}\thanks{\ \ Corresponding author.}, \textbf{Chang Liu}$^{2}$\footnotemark[1], {Tao Shen}$^{3}$,  {Can Xu}$^{2}$, \textbf{Xiubo Geng}$^{4}$, \\ \textbf{Binxing Jiao}$^{4}$ \and \textbf{Daxin Jiang}$^{4}$ \\
$^1$SKLSDE Lab, Beihang University \quad $^2$Peking University \\
$^3$AAII, FEIT, University of Technology Sydney  \quad $^4$Microsoft \\
 {\tt $^{1}$chongyang@buaa.edu.cn}\quad {\tt $^{2}$\{changliu,canxu\}@pku.edu.cn} \\
  {\tt $^{3}$tao.shen@uts.edu.cn} \quad
 {\tt $^{4}$\{xigeng,bxjiao,djiang\}@microsoft.com} \\
}

\begin{document}
\maketitle

\begin{abstract}
To improve the performance of the dual-encoder retriever, one effective approach is knowledge distillation from the cross-encoder ranker. 
Existing works prepare training instances by pairing each query with one positive and a batch of negatives. However,  most hard negatives mined by advanced dense retrieval methods are still too trivial for the teacher to distinguish, preventing the teacher from transferring abundant \emph{dark knowledge} to the student through its soft label.
To alleviate this issue, we propose \textsc{Adam}, a knowledge distillation framework that can better transfer the dark knowledge held in the teacher with \textsc{a}daptive \textsc{d}ark ex\textsc{am}ples.  Different from previous works that only rely on one positive and hard negatives as candidate passages, we create dark examples that all have moderate relevance to the query by strengthening negatives and masking positives in the discrete space. Furthermore, as the quality of knowledge held in different training instances varies as measured by the teacher's confidence score, we propose a self-paced distillation strategy that adaptively concentrates on a subset of high-quality instances to conduct our dark-example-based knowledge distillation to help the student learn better. We conduct experiments on two widely-used benchmarks and verify the effectiveness of our method.

\end{abstract}

\section{Introduction}

Information retrieval (IR) that aims to identify relevant passages for a given query is an important topic for both academic and industrial areas, and has powered many downstream tasks such as open-domain QA~\cite{chen17openqa} and knowledge-grounded conversation~\cite{dinan2018wizard}. 
Typically, IR systems usually follow the retrieve-and-re-rank paradigm \cite{hostatter2020improving, Huang2020Embedding,Zou2021Baidu} where a fast retriever first retrieved a bundle of relevant passages from a large-scale corpus through pre-built indices and then a more sophisticated ranker comes to re-rank these candidate passages to further obtain more accurate retrieval results.

\begin{figure}[t]
\centering
\includegraphics[width=0.8\columnwidth]{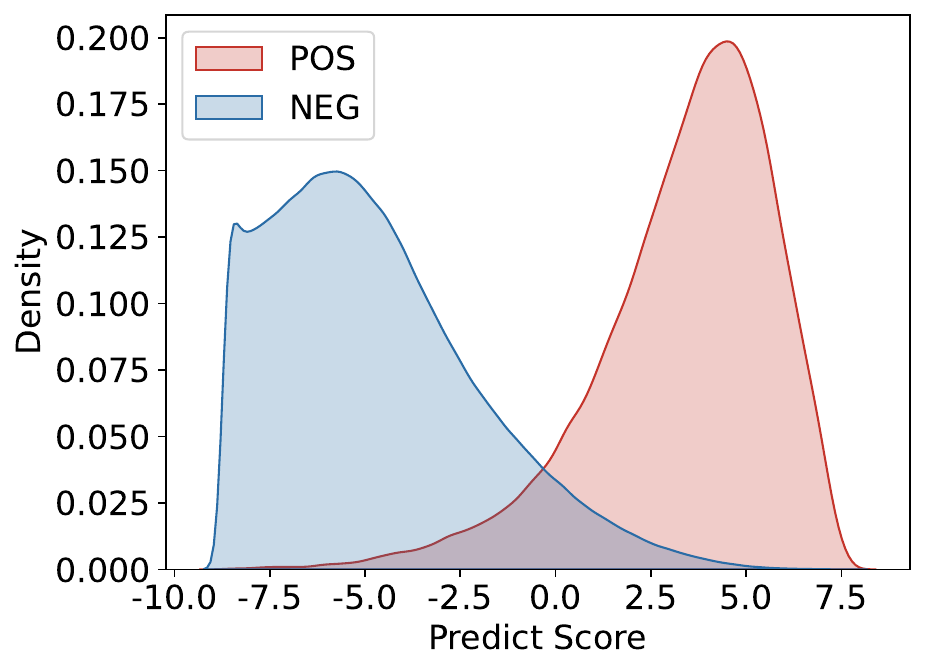}
\caption{Distributions of the prediction for the cross-encoder of R$^2$anker~\cite{zhou2022towards} over MS-MARCO. POS and NEG mean the distribution of positive and hard negatives respectively. The hard negatives are provided by RocketQAv2~\cite{Ren2021RocketQAv2}.}
\label{fig:dist}
\end{figure}

Under this paradigm, recent years have witnessed a growing number of works that utilize pre-trained language models (PLMs)~\cite{Qu2021RocketQA,Gao2021coCondenser} as retrievers and rankers to build IR systems. Among these efforts, there are two commonly adopted architectures: cross-encoder~\cite{bert2019naacl} that measure the relevance of a query-passage pair through jointly modeling their deep interactions; dual-encoder~\cite{Karpukhin2020DPR,Qu2021RocketQA} that encodes queries and passages separately into dense representations and calculate the similarity. Although dual-encoders are efficient for billions of indices, they suffer from inferior performance compared with cross-encoders since they can't capture the fine-grained semantic relevance between the query and the passage due to the absence of their deep interactions~\cite{Luan2021MEBERT}.
To help dual-encoders achieve better retrieval performance, a common practice is to draw on the powerful but cumbersome cross-encoder through knowledge distillation~\cite{google2020augmentation,Zhang2021AR2, Ren2021RocketQAv2, zeng2022curriculum, lin2023prod}. 
Along this line of research, various techniques are proposed to improve the knowledge transfer including data curriculum~\cite{lin2023prod,zeng2022curriculum}, on-the-fly distillation~\cite{Zhang2021AR2,Ren2021RocketQAv2} and new distillation objectives~\cite{lu2022ernie,menon2022defense}.

Though effective, we argue that existing dense retrieval distillation methods may not fully exploit the dark knowledge deeply held by the teacher.
In knowledge distillation~\cite{xu2018interpreting,lin2023prod}, the student learns not just the highest-scored class from the soft labels provided by the teacher, but also the entire probability distribution over classes, as this contains comprehensive fine-grained information referred to as "dark knowledge".
However, we empirically find that for existing distillation methods, the soft labels (i.e., the probability distributions over one positive and multiple negatives for a query) given by the teacher are too ``sharp'', despite they already adopted hard negatives~\cite{Ren2021RocketQAv2}.
As illustrated in Figure ~\ref{fig:dist}, we draw the score distributions of the positive and negative pairs using a pre-trained cross-encoder teacher.
It can be observed that the scores for most hard negatives are quite low (concentrated in (-7.5, -2.5)) and distributed far from the positives that have high scores. 
A similar observation is also drawn by~\citet{menon2022defense}.
This phenomenon indicates that even the hard negatives mined by the dense retriever are still too trivial for a well-trained cross-encoder teacher to distinguish, losing most of the utile dark knowledge.

To alleviate this issue, we propose ADAM, a knowledge distillation framework that can better exploit dark knowledge deeply held in the teacher by distillation with adaptive dark examples.
Our method originated from the intuition that 
a good soft label for the retriever to learn should be more {smooth}, which implies that the provided query-passage pairs should diversely distribute from highly-relevant pairs to loosely-relevant pairs from the view of the teacher.
To fill the gap between highly-relevant pairs and loosely-relevant pairs existing in current negative sampling methods, we propose two approaches to construct dark examples that all have moderate relevance to the query.
The first approach is to make negatives more relevant to the query by strengthening the negatives with the positive passage. 
The second approach is to make positives less relevant to the query by replacing some randomly selected tokens with mask tokens.
Considering that the newly created passages have moderate relevance to the query, we believe it is more appropriate to call them dark examples instead of negatives.
With these dark examples added, we successfully make the score distribution smoother as shown in Figure~\ref{fig:dist_new}, so that we can transfer more useful dark knowledge from the teacher. 
Moreover, since the soft label for different query-positive-negatives have different ``sharpness'' which we consider as an indication of how well the dark knowledge has been exploited, we further propose a self-paced distillation strategy that adaptively selects those examples whose soft labels are sharp to conduct our dark-example-based distillation to better transfer the dark knowledge.

We conduct experiments on two benchmarks, including MS-MARCO~\cite{Nguyen2016MSMARCO} and TREC Deep Learning 2019~\cite{Craswell2020TREC19}. In both benchmarks, the model is required to select the best response from a candidate pool. Evaluation results indicate that our method is significantly better than existing models on two benchmarks. To sum up, our contributions is three-fold:
\begin{itemize} 
    \item Propose to augment dark examples including reinforced negatives and noisy positives for more effective knowledge distillation in IR;
    \item Propose to adaptively concentrate on high-confidence training instances to better transfer knowledge;
    \item Empirical verify of the effectiveness of the proposed approach on two public datasets.
\end{itemize}
\section{Related Works}
There are two lines of research related to our work: dense retriever and knowledge distillation.

\paragraph{Dense Retriever.}
To overcome the vocabulary and semantic mismatch problems existing in conventional term-based approaches such as BM25~\cite{Robertson2009BM25}, researchers began to build neural retrievers upon pre-trained language models~\cite{Devlin2019BERT,Liu2019RoBERTa}.
In this way, the whole input text can be represented as a dense vector in a low-dimensional space (e.g., 768) and efficient retrieval can be achieved by approximate nearest neighbor search (ANN) algorithms such as FAISS~\cite{johnson2019billion}.
To learn a good dense retriever, various attempts have been made including hard negative mining~\cite{Karpukhin2020DPR,Luan2021MEBERT,Qu2021RocketQA,Xiong2021ANCE,Zhan2021STAR-ADORE}, retrieval-oriented pre-training~\cite{Lee2019ICT,Gao2021Condenser,Gao2021coCondenser}, knowledge distillation~\cite{Ren2021RocketQAv2,Zhang2021AR2,lu2022ernie,zhang2023led}, etc. We mainly focus on knowledge distillation in this paper.

\paragraph{Knowledge Distillation.}
Knowledge distillation~\cite{Hinton2015distilling,xu2024survey} aims to transfer the knowledge from a powerful teacher model to a student model to help it learn better. 
To achieve this goal, the student model is provided with the teacher's outputs as the supervision signal that it is enforced to mimic.
There are multiple types of supervision signals for the student to learn, including the teacher's output logits~\cite{Hinton2015distilling}, intermediate representations~\cite{romero2014fitnets}, relations of representations~\cite{park2019relational}, etc.
In the context of dense retrieval distillation, researchers basically adopt the cross-encoder as the teacher and use the teacher's probability distribution over candidate passages as the supervision signal. 
On this basis, several studies~\cite{Ren2021RocketQAv2,Zhang2021AR2,lu2022ernie} explored on-the-fly distillation to jointly optimize the teacher and the student, \citet{zeng2022curriculum} and \citet{lin2023prod} combined knowledge distillation with curriculum strategies to gradually improve the student.
Different from existing work, we focus on the quality of knowledge held in the teacher's soft label and propose to distill with adaptive dark examples to better transfer the dark knowledge to the student.

\section{Methodology}
In this section, we first introduce the preliminaries in dense retrieval distillation, then present our dark example augmentation method and adaptive distillation with dynamic data selection.

\subsection{Preliminary}
\label{sec:preliminary}

\paragraph{Task Description}
In this work, we study the learning of the dense retriever following the general setting of dense retrieval in existing work~\cite{Qu2021RocketQA,Ren2021RocketQAv2,Zhang2021AR2}.
Formally, there is a training set $\mathcal{D} = \{(q_i, \mathbb{P}_i)\}_{i=1}^{n}$ where $q_i$ is the query and $\mathbb{P}_i$ is the set of candidate passages. 
Commonly, $\mathbb{P}_i$ consists of a positive passage  $p^{+}_{i}$ and $m$ negative passages $\mathbb{P}^{-}_i = \{p^{-}_{i,j}\}_{j=1}^{m}$ constructed by random negative sampling~\cite{henderson2017efficient,gillick2018end} or hard negative mining~\cite{ANCE,Karpukhin2020DPR,Qu2021RocketQA}.
Based on $\mathcal{D}$, we aim to learn a retriever that can select the most relevant passage from the whole candidate pool.

\paragraph{Dual-Encoders}
A typical text retrieval system adopts the retrieve-and-rank paradigm, where the retriever is responsible for collecting a bubble of candidate passages and the ranker further re-ranks them.
Considering the trade-off between efficiency and accuracy, dual-encoders~\cite{Karpukhin2020DPR,Qu2021RocketQA,cai2022hyper} are often chosen as the retriever while cross-encoders~\cite{Devlin2019BERT} are usually adopted as the ranker.\footnote{We will use retriever and dual-encoder interchangeably.}

The dual-encoder-based retriever $\texttt{Enc}_\texttt{de}$ is responsible for encoding the given query $q_i$ and each of the candidate passage $p_j$ into dense vectors $\texttt{Enc}_\texttt{de}(q_i), \texttt{Enc}_\texttt{de}(p_j) \in \mathbb{R}^{h}$.
Then the relevance score for $q_i$ and $p_j$ is simply calculated as the inner product of their representations:
\begin{equation}\label{eq:de-score}
    \mathcal{R}_\texttt{de}(q_i, p_j) = \texttt{Enc}_\texttt{de}(q_i)^{\top} \cdot \texttt{Enc}_\texttt{de}(p_j).
\end{equation}
To fulfill this goal, the retriever is typically trained with supervised contrastive loss:
\begin{equation*}\label{eq:de-loss-sup} \small
    \mathcal{L}_\texttt{sup} = - \log \frac{\exp^{\mathcal{R}_\texttt{de}(q_i, p^{+}_{i})}}
    {\exp^{\mathcal{R}_\texttt{de}(q_i, p^{+}_{i})} + 
    \sum\limits_{p^{-}_{i,j} \in \mathbb{P}^{-}_{i}} \exp^{\mathcal{R}_\texttt{de}(q_i, p^{-}_{i,j})}}.
\end{equation*}
where $p_{i}^{+}$ is the labeled positive document paired with $q_i$ and $\mathbb{P}^{-}_{i}$ denotes the set of candidate documents for $q_i$ which is typically constructed during training by random negative sampling or hard negative mining methods.

\paragraph{Cross-Encoders}
The cross-encoder ranker $\texttt{Enc}_\texttt{ce}$ is in charge of calculating the matching score of $q_i$ and $p_j$ more accurately as it can model their fine-grained interactions, and re-ranking the retrieved candidate passages provided by the retriever to improve the retrieval results.
Concretely, given a query $q_i$ and a passage $p_j$, the input is formed as the concatenation of $q$ and $p$ with $[\mathtt{CLS}]$ in the beginning and $[\mathtt{SEP}]$ as their separation and is fed into transformer~\cite{vaswani2017attention}. The representation of $[\mathtt{CLS}]$ in the top layer is used to calculate the relevance score with a projection head $f(\cdot)$:
\begin{equation}\label{eq:ce-score}
    \mathcal{R}_\texttt{ce}(q_i, p_j) = f(\texttt{Enc}_\texttt{ce}([\mathtt{CLS}], q_i, [\mathtt{SEP}], p_j)).
\end{equation}

\paragraph{Knowledge Distillation in IR}
As cross-encoders are more capable of measuring the relevance of $q_i$ and $p_j$ than dual-encoders but at a cost of computational inefficiency, it's promising to transfer the knowledge from the strong cross-encoders to the weak dual-encoders through knowledge distillation~\cite{Zhang2021AR2, Ren2021RocketQAv2,zeng2022curriculum,lu2022ernie,lin2023prod}. 
In dense retrieval distillation, as both the positive passage $p^{+}_i$ and the negatives $\mathbb{P}^{-}_{i}$ can be treated uniformly, we use $\mathbb{P}_{i} = \{p^{+}_i\} \cup \mathbb{P}^{-}_{i}$ to denote the whole candidate set of passages.
The relevance score of $q_i$ and each $p_j \in \mathbb{P}_{i}$ can be calculated using a dual-encoder $\texttt{Enc}_\texttt{de}$ and a cross encoder $\texttt{Enc}_\texttt{ce}$ using Eq.~\ref{eq:de-score} and Eq.~\ref{eq:ce-score}. Then, the probability distributions over candidate passages of the dual-encoder and the cross-encoder $\bm p_{de,i}, \bm p_{ce,i} \in \mathbb{R}^{|\mathbb{P}_{i}|}$ are calculated by normalizing the relevance scores over $\mathbb{P}_{i}$, where each element is calculated as:
\begin{equation}\label{eq:dece-distribution}
\begin{aligned}
\hat{\mathcal{R}}_{de, i}^{j} =  \frac{\exp^{\mathcal{R}_\texttt{de}(q_i, p_j)}}
{\sum_{p_k \in \mathbb{P}_{i}} e^{\mathcal{R}_\texttt{de}(q_i, p_k)}} \\
\hat{\mathcal{R}}_{ce, i}^{j} = \frac{\exp^{\mathcal{R}_\texttt{ce}(q_i, p_j)}}
{\sum_{p_k \in \mathbb{P}_{i}} \exp^{\mathcal{R}_\texttt{ce}(q_i, p_k)}}.
\end{aligned}
\end{equation}
To distill the knowledge from the cross-encoder to the dual-encoder, the distribution of the cross-encoder $\hat{\mathcal{R}}_{ce, i}$ is considered as the soft label that guides the learning of the dual-encoder by minimizing the KL-divergence between $\hat{\mathcal{R}}_{ce, i}$ and $\hat{\mathcal{R}}_{de, i}$:
\begin{equation}\label{eq:de-loss-kd}
    \mathcal{L}_{kd} = - \sum_{ (q_i, \mathbb{P}_i) \in \mathcal{D}} \text{KL-Div}(\hat{\mathcal{R}}_{ce, i} || \hat{\mathcal{R}}_{de, i})
\end{equation}

\begin{figure}[t]
\centering
\includegraphics[width=0.8\linewidth]{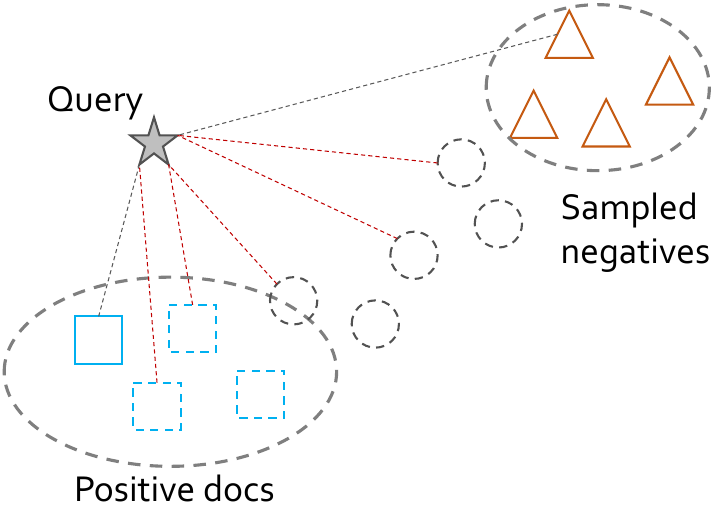}
\caption{Illustration of dark examples. The solid rectangle and triangles mean the gold passage and the negative passages respectively. Dotted rectangles and circles denote noisy positives and mixed samples respectively.}
\label{fig:darkexamples}
\end{figure}

\subsection{Dark Examples Construction}\label{sec:construction}
When transferring the knowledge from the cross-encoder teacher to the dual-encoder student using Eq.~\ref{eq:de-loss-kd}, the set of candidate passages $\mathbb{P}_{i}$ plays a vital role.
Previous works in dense retrieval distillation~\cite{Zhang2021AR2, Ren2021RocketQAv2, zeng2022curriculum, lu2022ernie,lin2023prod} simply follow the supervised learning setting where they utilize $\mathbb{P}_{i} = \{p^{+}_i\} \cup \mathbb{P}^{-}_{i}$ as the candidate set.
However, by empirical analyses on Fig.~\ref{fig:dist}, we have found that the negative set $\mathbb{P}^{-}_{i}$ produced by existing hard negative mining approaches~\cite{Qu2021RocketQA} is too trivial for the cross-encoder teacher, which makes the soft label provided by the cross-encoder teacher too sharp at the positive passage and therefore prevents the student from learning utile dark knowledge hidden in the distribution of other passages (i.e.,  negatives).

We suppose smoother soft labels naturally obtained (instead of scaled by softmax temperature) can be better knowledge carriers that transfer the dark knowledge. 
Given the teacher and the query, we point out that the natural way to smoothen the soft label is to operate on the set of candidate passages, or more precisely, to replace the original set of candidate passages $\mathbb{P}_{i}$ that are either too relevant or too irrelevant from the teacher's view with new ones 
$\tilde{\mathbb{P}}_{i}$ whose relevance to the query cannot be easily tell apart by the cross-encoder teacher.

To construct the new set of candidate passages that satisfy this desired characteristic, we propose two dual approaches that operate on the original positive passage $p^{+}_{i}$ and the negative set $\mathbb{P}^{-}_{i}$ respectively.
We name the newly constructed passages in $\tilde{\mathbb{P}}_{i}$ dark examples to demonstrate that can no longer be simply categorized into positives and negatives as they have moderate relevance to the query.
An illustration of dark examples is shown in Figure~\ref{fig:darkexamples}.
It should be noticed that it is the specific setting of knowledge distillation where the supervision signal is derived from the teacher's soft label instead of human labels that make it possible to learn from dark examples.

\paragraph{Sampled Negatives.}
Early works~\cite{henderson2017efficient, gillick2018end} randomly choose negative passages by considering the passages of other query-passage pairs within the same mini-batch as the negatives. 
More recently, researchers use BM25~\cite{Karpukhin2020DPR} or dual-encoders~\cite{ANCE} to select hard negatives globally from the whole candidate passages with the fast retrieval method~\cite{Qu2021RocketQA,Ren2021RocketQAv2}. 
{We will compare the effectiveness of random negatives (denoted as Rand) and hard negatives (denoted as Hard) with our method (denoted as Dark) in experiments.}

\paragraph{Dark Examples with Reinforced Negatives}
The reasonable way to create dark examples based on $\mathbb{P}^{-}_{i}$ is to make hard negatives harder, or in other words, more relevant to the query.
To achieve this goal, it is non-trivial to accurately edit the semantics of a negative passage towards increasing its relevance to the query with controllable text generation techniques. 
Instead, we propose a rather simple yet effective approach that mixes up query-relevant content with negative passages to directedly strength their relevance to the query. 
Based on this motivation, we consider mixing up hard negatives with the positive passage\footnote{We also tried to make the hard negatives even harder by mixing up hard negatives with the query following~\citet{kalantidis2020hard}, however, we found little change in performance.}. 
Formally, given a training example $(q_i, p^{+}_i, \mathbb{P}^{-}_i)$, we concatenate $p^{+}_i$ with each of the negative passage $p^{-}_{i,j}$ to form the set of dark examples for $q_i$:
\begin{equation}
\begin{aligned}
\label{eq:mix}
   \mathcal{N}^{rein}_{i} = \{ p^{+}_{i} [\mathtt{SEP}] p^{-}_{i,j} \}_{j=1}^{m}.
\end{aligned}
\end{equation}
Here, we choose to mix-up passages at the lexical level instead of the embedding space~\cite{guo2019augmenting} because our method can produce valid language inputs and can preserve the relevant cues while introducing some less-relevant content.
We also tried mixing-up negatives with the positive in the embedding space but found this kind of mix-up resulted in low-quality predictions of the cross-encoder teacher since it has never seen samples based on mixed embeddings during training.



\paragraph{Dark Examples with Noisy Positive}
Different from the above approach that creates dark examples by making hard negatives harder, we also consider the opposite direction: making the positive passage $p^{+}$ not that relevant to the query by introducing noise. 
We achieve this goal by input-masking~\cite{Devlin2019BERT}. 
Given the positive passage $p^{+}_i$ for the query $q_i$, we randomly sample a subset of tokens from $p^{+}_i$ and replace them with the special token $[\mathtt{MASK}$] with the masking ratio $m_r$:
\begin{equation}
\begin{aligned}
\label{eq:mask}
   \mathcal{N}^{mask}_i = \{ \mathtt{MASK}_{m_r}(p^{+}_i) \}_{m_{r}}.
\end{aligned}
\end{equation}
To generate noisy positives with more diverse relevant to the query, we use masking with a variety of masking ratios.

\subsection{Distillation with Adaptive Dark Examples}\label{sec:adaptive}
We have elaborated our motivation and approach to create dark examples, the remaining question is how to conduct effective knowledge distillation with dark examples. 
Existing knowledge distillation methods using all the labeled data without distinction, which we argue is sub-optimal.
As knowledge distillation relies on the teacher's prediction as the supervision signal, the ``quality'' of knowledge held in the teacher's soft label naturally varies among different training examples. 
We assume that those training examples that the teacher is more confident than others are better carriers of knowledge for three reasons:
(1) These instances are far from the decision boundaries of the model, and thus the corresponding passages are more likely to be true positives and true negatives, avoiding data noise.
(2) Only the knowledge held in the instances that the teacher can cope with well are reliable and worth to be learned by the student.
(3) The teacher's soft label for the high-confidence instances is too sharp, which indicates the dark knowledge held in these reliable instances has not been well exploited.

Therefore, we propose to adaptively concentrate on these high-confidence training instances during the training process to conduct our dark-example-based knowledge distillation.
Formally, for a training instance, we can calculate the log-probability of the positive passage $p^{+}_i$ against negatives $\mathbb{P}^{-}_{i}$ with the teacher as the confidence score:
\begin{equation} 
\label{eq:teacher-confidence}
    \mathcal{C}(q_i) = \log \frac{\exp^{\mathcal{R}_{ce}(q_i, p^{+}_i)}}
    {\exp^{\mathcal{R}_{ce}(q_i, p^{+}_i)} + 
    \sum\limits_{p^{-}_{i,j} \in \mathbb{P}^{-}_{i}} \exp^{\mathcal{R}_{ce}(q_i, p^{-}_{i,j})}}.
\end{equation}

Suppose the training process consists of $T$ epochs, in each epoch $t$,
 we can sort a batch of training instances $\mathcal{B}_t$ in ascending order based on the confidence scores. Then we adaptively select the subset of instances $\hat{\mathcal{B}}_t$ in the batch that have the highest confidence scores with the ratio $(1-\frac{t}{2*T})$ to construct dark examples:
\begin{equation}
\begin{aligned}
\tilde{\mathcal{B}}_{t} & =\mathop{\arg\max}_{q_i \in \mathcal{B}_t, \tilde{\mathcal{B}}_{t}\subset \mathcal{B}_t, \|\tilde{\mathcal{B}}_{t}\|={(1-\frac{t}{2*T}) \times b}} \mathcal{C} (q_i).
\label{eq:select}
\end{aligned}
\end{equation}
where $b$ is the batch size for training.


Thereby, we have two sets in each step of the $t$-th training epoch: the original training batch $\mathcal{B}_t$ and the subset with the highest confidence that has both original candidate passages and our created dark examples $\tilde{\mathcal{B}}_{t}$.
We jointly optimize the student with the supervised loss (Eq.~\ref{eq:de-loss-sup}) on $\mathcal{B}_t$ and the knowledge distillation loss (Eq.~\ref{eq:de-loss-kd}) on $\tilde{\mathcal{B}}_{t}$:
\begin{equation}
\begin{aligned}
\mathcal{L}_{t} = \lambda \cdot & \sum_{\mathcal{B}_t \in \mathcal{D}}  \sum_{(q_i, \mathbb{P}_{i})\in \mathcal{B}_t}  \mathcal{L}_{sup} + \sum_{\hat{\mathcal{B}}_t \in \mathcal{D}} \sum_{(q_i, \tilde{\mathbb{P}}_{i})\in \tilde{\mathcal{B}}_t } \mathcal{L}_{kd}.
\end{aligned}
\end{equation}
where  $\tilde{\mathbb{P}}_{i} = \{\mathbb{P}^{-}_{i} \cup \mathcal{N}^{mix}_{i} \cup \mathcal{N}^{mask}_{i} \}$ is the new candidate set for $q_i$, and $\lambda$ is a hyper-parameter as a trade-off between the supervised objective and distillation objective with adaptive dark examples.

\section{Experiments}

\begin{table*}[ht!]
\resizebox{0.95\textwidth}{!}{
    \centering
    \begin{tabular}{llcccccc}
    \toprule
        \multirow{2}*{\textbf{Methods}} & \multirow{2}*{\textbf{PLM}}  & \multirow{2}*{\textbf{KD}} &  \multicolumn{3}{c}{\textbf{MS-MARCO Dev}} & \multicolumn{2}{c}{\textbf{TREC DL 19}} \\
              & & & MRR@10 & R@50 & R@1000 & NDCG@10  &R@100  \\
        \midrule
        \textbf{Sparse retrieval} \\
        BM25 (anserini)~\cite{bm25} & - & - & 18.7 & 59.2 & 85.7  & 50.6 & -\\

        doc2query~\cite{doc2query} & -  & -& 21.5 & 64.4 & 89.1 & - & -  \\
        DeepCT~\cite{deepct2019sigir} & BERT$_\text{base}$ & - & 24.3 & 69.0 & 91.0 & 55.1 & -  \\
        docTTTTTquery~\cite{doctttttquery} & - & - & 27.7 & 75.6 & 94.7 & - & -  \\
        UHD-BERT~\cite{uhdbert2021} & BERT$_\text{base}$ & - & 29.6 & 77.7 & 96.1 & - & - \\        
        COIL-full~\cite{gao2021coil} & BERT$_\text{base}$& - & 35.5 & - & 96.3 & 70.4 & -  \\
        UniCOIL~\cite{Lin2021UniCOIL} & BERT$_\text{base}$& -&35.2 & 80.7 & 95.8 & - & - \\
        SPLADE-max~\cite{Formal2021SPLADEv2} &  BERT$_\text{base}$ & -& 34.0 & - & 96.5 & 68.4 & - \\
         Unifier$_{\text{lexicon}}$~\cite{shen2023unifier} & coCon$_\text{base}$ & $\checkmark$ &  39.7 & - &  98.1 & 73.3 & - \\
         
        \midrule \midrule
        \textbf{Dense retrieval} \\
        DPR-E~\cite{Ren2021RocketQAv2} & ERNIE$_\text{base}$ & -& 32.5 & 82.2 & 97.3 &  - &  -\\
        ANCE (single)~\cite{ANCE} & RoBERTa$_\text{base}$ & -& 33.0 & - & 95.9 & 65.4  & 44.5 \\
        TAS-Balanced~\cite{tas2021} & BERT$_\text{base}$ & $\checkmark$& 34.0 & - & -  & 71.2& - \\
        ME-BERT~\cite{MEBERT} & BERT$_\text{large}$ & -& 34.3 & - & - & - & -  \\
        ColBERT~\cite{colbert2020sigir} & BERT$_\text{base}$& - & 36.0 & 82.9 & 96.8& 67.0  & -  \\
        ColBERT v2~\cite{Khattab2021ColBERTv2} & BERT$_\text{base}$ & $\checkmark$& 39.7 & 86.8 & 98.4  & 72.0 & - \\

        ADORE+STAR~\cite{Optimizing2021sigir} & RoBERTa$_\text{base}$ & -& 34.7 & - & -  & 68.3 & -\\
        Condenser~\cite{Gao2021Condenser} &  BERT$_\text{base}$& - & 36.6 & - & 97.4 & - & -  \\
        RocketQA~\cite{Qu2021RocketQA} & ERNIE$_\text{base}$ & -& 37.0 & 85.5 & 97.9 & -  & - \\
        PAIR~\cite{pair} & ERNIE$_\text{base}$ & -& {37.9} & {86.4} & {98.2} & - & - \\
        CoCondenser~\cite{gao2022unsupervised} &  BERT$_\text{base}$ & -& 38.2 & - & 98.4 & - & - \\
        RocketQAV2~\cite{Ren2021RocketQAv2} &  BERT$_\text{base}$ & $\checkmark$&  38.8 & 86.2 & 98.1 & - & -\\
        AR2~\cite{Zhang2021AR2} &  BERT$_\text{base}$ & $\checkmark$& 39.5 & - & 98.6 & - & -  \\
        CL-DRD~\cite{zeng2022curriculum} & DistilBERT & $\checkmark$&  38.2 &  - &  -  & 72.5 & 45.3\\
        ERNIE-Search~\cite{lu2022ernie}   & BERT$_\text{base}$& $\checkmark$ & 40.1 & 87.7 & 98.2 &- & - \\
        RetroMAE~\cite{xiao2022retromae} & BERT$_\text{base}$ & $\checkmark$ &  39.3 & 87.0 & 98.5 & - & - \\

        Unifier$_{\text{dense}}$~\cite{shen2023unifier} & coCon$_\text{base}$ & $\checkmark$ & 38.8 & - &  97.6 &  71.1 & -\\
        
        bi-SimLM~\cite{wang2022simlm} & BERT$_\text{base}$ & $\checkmark$ & 39.1 & 87.3 & 98.6 & 69.8 & -\\
        PROD~\cite{lin2023prod} & ERNIE-2.0-BASE & $\checkmark$ & 39.3 & 87.1 & 98.4 & 73.3 & 48.4\\
        InDi~\cite{cohen2024indi} & coCon$_\text{base}$ & - &  38.8 & 86.6 & 98.5 & - & -\\
        \midrule
        Rand KD (\emph{Teacher = RocketQAV2}) & BERT$_\text{base}$ & $\checkmark$ &  38.1  & 86.9 & 98.2 & - & -\\
        Hard KD (\emph{Teacher = RocketQAV2}) & BERT$_\text{base}$ & $\checkmark$ & {39.1} & {87.6} & {98.5} &  - & - \\ 
        \rowcolor{LightCyan} \modelname \ \ (\emph{Teacher = RocketQAV2})  & BERT$_\text{base}$ & $\checkmark$& {39.8} & {88.1} & {98.6} &  72.1 & 50.3 \\ 
        \midrule
        Rand KD  (\emph{Teacher = R$^2$anker}) & BERT$_\text{base}$ & $\checkmark$& 38.1 & 86.0 & 97.9 & {-} & - \\
        Hard KD  (\emph{Teacher = R$^2$anker})  & BERT$_\text{base}$ & $\checkmark$& {40.0} & 87.6  & 98.1 & - & -\\
        \rowcolor{LightCyan} \modelname \ \ (\emph{Teacher = R$^2$anker}) & BERT$_\text{base}$& $\checkmark$ & \underline{41.0} & \underline{88.5} & {98.5} &  \underline{73.4}  &   \underline{49.8} \\ 
    \bottomrule
    \end{tabular}
    }
    \caption{Passage retrieval results on MS-MARCO and TREC DL 19 datasets. PLM is the abbreviation of the pre-trained language Model. KD indicates whether a model is distilled by a ranker. We copy the results from original papers and leave them blank if the original paper does not report the result. The best results are in underlined fonts.}
    \label{tab:main_results}
\end{table*}

We evaluate our method on two public human-annotated real-world benchmarks, namely MS-Marco and TREC Deep Learning 2019.

\subsection{Datasets and Evaluation Metrics}
Consisting with previous studies on dense information retrieval~\cite{Hofstatter2021TAS-B,Xiong2021ANCE}, we use popular passage retrieval datasets,
MS-MARCO~\cite{Nguyen2016MSMARCO}. The dataset contains 8.8M passages from Web
pages gathered from Bing’s results to real-world
queries.  
The training set contains about
500k pairs of query and relevant passage, and the dev set consists of $6,980$ queries. Based on the queries and passages in the dataset, MS-MARCO passage retrieval and ranking tasks were created. Following previous works~\cite{zeng2022curriculum}, we report the performance on MS-MARCO Dev set as well as TREC Deep Learning (DL) 2019 set \cite{Craswell2020TREC19} which includes 43 queries. We report MRR@10
and Recall@50/1K for MS-MARCO, and nDCG@10 and Recall@100 for TREC DL 19.
We also report zero-shot transfer performance (nDCG@10) on BEIR benchmark~\cite{Thakur2021BEIR}.

\subsection{Baselines}
To make a comprehensive comparison, we choose the following state-of-the-art approaches as baselines. These methods contain both sparse and dense passage retrievers.

The sparse retrieval methods include the traditional retriever BM25~\cite{Yang2017Anserini} and several representative sparse retrievers, including doc2query~\cite{negative2020google}, DeepCT~\cite{Dai2019DeepCT}, docTTTTT-query~\cite{doctttttquery}, UHD-BERT~\cite{uhdbert2021}, COIL-full~\cite{gao2021coil}, UniCOIL~\cite{Lin2021UniCOIL}, and SPLADE-max~\cite{Formal2021SPLADEv2}.

The dense retrieval methods produce continuous neural vectors for each passage and query. The methods include DPR-E~\cite{Qu2021RocketQA}, ANCE~\cite{Xiong2021ANCE},  TAS-Balanced~\cite{Hofstatter2021TAS-B}, ME-BERT~\cite{Luan2021MEBERT}, ColBERT~\cite{Khattab2020ColBERT}, ColBERT v2~\cite{Khattab2021ColBERTv2}, NPRINC~\cite{Lu2021SeedEncoder},  ADORE+STAR~\cite{Zhan2021STAR-ADORE}, Condenser~\cite{Gao2021Condenser}, RocketQA~\cite{Qu2021RocketQA}, PAIR~\cite{Ren2021PAIR}, CoCondenser~\cite{gao2022unsupervised}, RoketQAV2~\cite{Ren2021RocketQAv2}, AR2~\cite{Zhang2021AR2}, CL-DRD~\cite{zeng2022curriculum}, ERNIE-Search~\cite{lu2022ernie}, RetroMAE~\cite{xiao2022retromae}, Unifier~\cite{shen2023unifier}, bi-SimLM~\cite{wang2022simlm}, PROD~\cite{lin2023prod} and InDi~\cite{cohen2024indi}. Some of them are enhanced by knowledge distillation from the ranker. For example, RoketQAV2, AR2, and ERNIE-Search introduce the on-the-fly distillation method. CL-DRD and PROD propose progressive distillation with a data curriculum to gradually improve the student.

\subsection{Implementation Details}
Consisting with the setting of RocketQA V2~\cite{Ren2021RocketQAv2}, we choose the learned dual-encoder in the first step of RocketQA~\cite{Qu2021RocketQA} as the initialization of our dense retriever\footnote{The retriever can also be replaced with other trained retriever. We observed that using the trained model to initialize the retriever can help achieve slightly better results.}.
We adopt two advanced cross-encoder rankers as our teacher model: RocketQAV2~\cite{Ren2021RocketQAv2} and R$^2$anker~\cite{zhou2022towards}\footnote{The results of BM25-reranking on MS-MARCO Dev for R$^2$anker~\cite{zhou2022towards} and RocketQAV2~\cite{Ren2021RocketQAv2} are 40.1 and 40.7 respectively.}. 
We randomly select $m$ hard negatives provided by~\citet{Ren2021RocketQAv2} for each query. 
For supervised learning, a positive passage and all the selected negatives are used.
While for distillation, the candidate passage set for a query consists of $m$ original negatives, $m$ dark examples in $\mathcal{N}^{mix}_{i}$, and $5$ dark examples in $\mathcal{N}^{mask}_{i}$ with different masking ratios $m_{r} \in \{0.15,0.25,0.35, 0.45, 0.55\}$.
We set the number of negatives $m$ to $10$ from $\{5,10,15,20,25,30\}$\footnote{We found $m=15$ to be the optimal parameter. However, considering that our method will expand the number of negatives with the augmented dark examples, we set m=10 in our experiment.}.
We set the maximum lengths for queries and passages as $32$ and $128$. The dropout rate is set to $0.1$ on the cross-encoder. 
In training, we use AdamW~\cite{Loshchilov2017FixingWD} as the optimizer to train the model. We set the batch size as $128$, the peak learning rate as $5e-5$, and the warm-up steps as $100$. 
We set the weight $\lambda$ for the supervised objective as $0.01$ by varying it in $\{0.001,0.01,0.05, 0.1, 0.5\}$. 

\subsection{Overall Performance}
We report the overall evaluation results on MS-MARCO and TREC Deep Learning 2019 respectively.
On both benchmarks, we not only show the performance of our dual-encoder retriever under knowledge distillation from two different cross-encoder teachers, but also provide comparisons between different choices of construction of candidate set $\mathbb{P}_i$.
The main results are shown in Table~\ref{tab:main_results}. We can draw three main conclusions:

{\textit{Our created dark examples improve the performance of knowledge distillation over hard negatives and random negatives.}}
With the same cross-encoder as the teacher, we analyze the impact of how the candidate set of passages is constructed.
It can be observed that using random negatives results in poor performance and the integration of hard negative mining indeed improve the performance.
When equipped with our created dark examples which are even harder than existing hard negatives, our model further makes a substantial improvement over that using hard negatives.

{\textit{Our framework \textsc{Adam} is compatible with different teachers.}}
To test the generalization ability over different teachers, we conduct experiments using two advanced cross-encoders (R$^2$anker and RocketQAV2) as the teacher. 
Consistent improvement can be observed when using our proposed dark examples for knowledge distillation with the two different teachers.
Moreover, we can compare the effectiveness of the two teachers.
When using random negatives, knowledge distillation with the two teachers results in comparable results. 
But when using hard negatives and dark examples, the model distilled by R$^2$anker yields significantly better performance than its counterparts. 
Therefore, for the remaining ablation studies and analyses, we use R$^2$anker as the teacher by default.

{\textit{With R$^2$anker as the teacher, our method (the bottom line) achieves superior performance over most baselines.}} 
Our model achieves $41.00$ on MRR@10 on the development set of MS-MARCO, outperforming most of the existing methods and is comparable with SimLM~\cite{wang2022simlm} which is obtained by a time-consuming large-scale pre-training followed with a cumbersome multi-stage supervised fine-tuning.

\begin{table*}[t]
\small 
\centering
\setlength{\tabcolsep}{1.5pt}
\begin{tabular}{l!{\color{lightgray}\vrule}ccc!{\color{lightgray}\vrule}c!{\color{lightgray}\vrule}ccccccc}

\toprule
Rep type& \multicolumn{3}{c!{\color{lightgray}\vrule}}{Sparse} & Mul-vec& \multicolumn{7}{c!{\color{lightgray}}}{Dense }  \\  
\midrule

\textbf{Method} & BM25      & SPLADE    & UnifieR     & ColBERT & DPR   & ANCE      & TAS-B  & CoCond       & CL-DRD &RocQA 
 & \modelname\           \\ \midrule

 Distillation&  \xmark  & \cmark & \cmark &  \cmark &  \xmark& \xmark& \cmark& \xmark& \cmark& \cmark& \cmark\\
 \midrule
TREC-COVID                                            & 65.6             & 71.0  & 71.5 & 73.8     & 33.2  & 65.4     & 48.1          & 71.2         &58.4&67.5 &   {73.0} \\
NFCorpus                                              & 32.5      & 33.4 & {32.9} & 33.8   & 18.9  & 23.7  & 31.9          & 32.5            & 31.5& 29.3 & {31.5} \\
FiQA                                                  & 23.6        & 33.6  & 31.1  & {35.6}  & 11.2  & 29.5   & 30.0           & 27.6        &30.8&30.2        & 31.5 \\
ArguAna                                               & 31.5           & 47.9 &  39.0   & 46.3         & 17.5  & 41.5 & 42.9      & {29.9} &41.3& 45.1     &  40.3 \\
Tóuche-2020                                           & {36.7}   & 27.2  &  30.2  & 26.3   & 13.1  & 24.0  & 16.2        & 19.1          & 20.3&24.7    &  25.6 \\
Scidocs                                               & 15.8             & 15.8 & 15.0   & 15.4    & 7.7   & 12.2  & 14.9         & 13.7          & 14.6 & 13.1      &   14.1 \\
SciFact                                               & 66.5           & 69.3   &  {68.6}  & 69.3      & 31.8  & 50.7  & 64.3     & 61.5         & 62.1& 56.8 &  59.4  \\ 

NQ                                                    & 32.9             & 52.1 & 51.4  &     56.2     & 47.4  & 44.6    & 46.3    & 48.7       & 50.0 & 50.5    &  51.9  \\
HotpotQA                                              & 60.3        & 68.4 & {66.1}   & 66.7     & 39.1  & 45.6  & 58.4       & 56.3         & 58.9 & 53.3 & 58.6 \\
DBPedia                                               & 31.3        & 43.5   &  40.6  & {44.6}     & 26.3  & 28.1  & 38.4      & 36.3          & 38.1 & 35.6      &  39.6\\
Fever                                                 & 75.3         & 78.6  & 69.6    & 78.5     & 56.2  & 66.9     & 70.0     &  49.5           & 73.4& 67.6  &   66.8  \\
Climate-FEVER                                         & 21.3           & {23.5}  &  17.5    & 17.6       & 14.8  & 19.8   & 22.8      &  14.4       &20.4& 18.0    &  21.4  \\
\midrule
 AVERAGE                                               & 41.1         & 42.4      &   44.5  &  47.0   & 26.4 & 37.7      & 40.4       & 38.9 &  42.0 & 41.0 & 42.8 \\ 
\bottomrule
\end{tabular}
\caption{\small Zero-shot transfer performance (nDCG@10) on BEIR benchmark. `BEST ON' and `AVERAGE' do not take the in-domain result into account.  `ColBERT' is its v2 version \citep{Khattab2021ColBERTv2}.  `CoCond' refers to CoCondenser~\cite{Gao2021coCondenser} and `RocQA' means RocketQAV2~\cite{Ren2021RocketQAv2}.} \label{tab:exp_beir_details}

\label{tab:beir}
\end{table*}

\begin{table}[t!]
    \small
    \centering
    \begin{tabular}{lc}
    \toprule
        Methods &  MRR@10 \\ \midrule
        \modelname\ \  & \underline{38.99}  \\ 
        \midrule
        w/o. $\mathcal{N}^{rein}$ \ \ (Eq.\ref{eq:mix})  & 38.82   \\ 
        w/o. $\mathcal{N}^{mask}$ (Eq.\ref{eq:mask})  & 38.76   \\ 
        w/o. \{$\mathcal{N}^{rein}$ \& $\mathcal{N}^{mask}$ \} & 38.64  \\ 
        w/o. \{$\mathcal{N}^{rein}$ \& $\mathcal{N}^{mask}$ \& ADA \}  & {38.61} \\ 
        w/o. \{$\mathcal{N}^{rein}$ \& $\mathcal{N}^{mask}$ \& ADA \& $\mathcal{L}_\texttt{sup}$\}  & {38.36}  \\ 
    \bottomrule
    \end{tabular}
    \caption{Ablation results on MS-Marco. We report the reranking performance.}
    \label{tab:ablation}
\end{table}

\subsection{Ablation study}

We have analyzed the overall performance on two benchmarks and proved the effectiveness of our method. 
Here, we conduct ablation studies to verify the indispensability of each crucial design.
We provide the results of the ablation study in Table~\ref{tab:ablation}.

{\textit{Dark examples.}}
Recall that we propose two types of methods to construct dark examples: (1) strengthening negatives ($\mathcal{N}^{rein}$) by mixing with the positive to make negatives more relevant to the query, and (2) polluting positives (($\mathcal{N}^{mask}$)) to make positives not that relevant. 
We first test the individual effect of $\mathcal{N}^{rein}$ and $\mathcal{N}^{mask}$. When removing each of them individually, performance drops can be observed. And when we remove both of them, the model performs worse. This observation indicates that the incorporation of both $\mathcal{N}^{rein}$ and $\mathcal{N}^{mask}$ is beneficial to the overall performance.

{\textit{Distillation with adaptive dark examples.}}
In addition to dark examples, we also introduce a self-paced distillation algorithm that can better transfer dark knowledge with adaptive dark examples. 
When this strategy is removed, we create dark examples for all the training instances.
It can be seen that distillation adaptively using the subset of instances that the teacher is most confident is better than using the whole training set, which is in accord with our assumption that the instances with higher confidence are a better carrier of knowledge.

\textit{Distillation with additional supervised loss.}
Although the teacher's soft label provides abundant dark knowledge for the student to learn, we also involve the traditional supervised loss.
We can observe that although the weight $\lambda$ for supervised loss is quite small (i.e., 0.01), we find this term indispensable for the overall performance.

\subsection{Discussions}

\vspace{1mm}
{\textit{Zero-shot transfer performance}}. We also curious about the transfer ability of our method and conduct experiments on the BEIR benchmark. Table~\ref{tab:beir} reports the zero-shot performance. We can find that our method surpasses existing dense approaches and three distillation methods (TAS-B, CL-DRD, and RocketQAV2). These results demonstrate that the reranker can transfer knowledge more effectively to the dense retriever using dark examples, and its out-of-distribution (OOD) adaptation ability is also well inherited by the retriever.

\begin{figure}[t!]
  \centering
  \subfigure[{The impact of $m$}] { \label{fig:trend_m}
    \includegraphics[width=0.476\columnwidth]{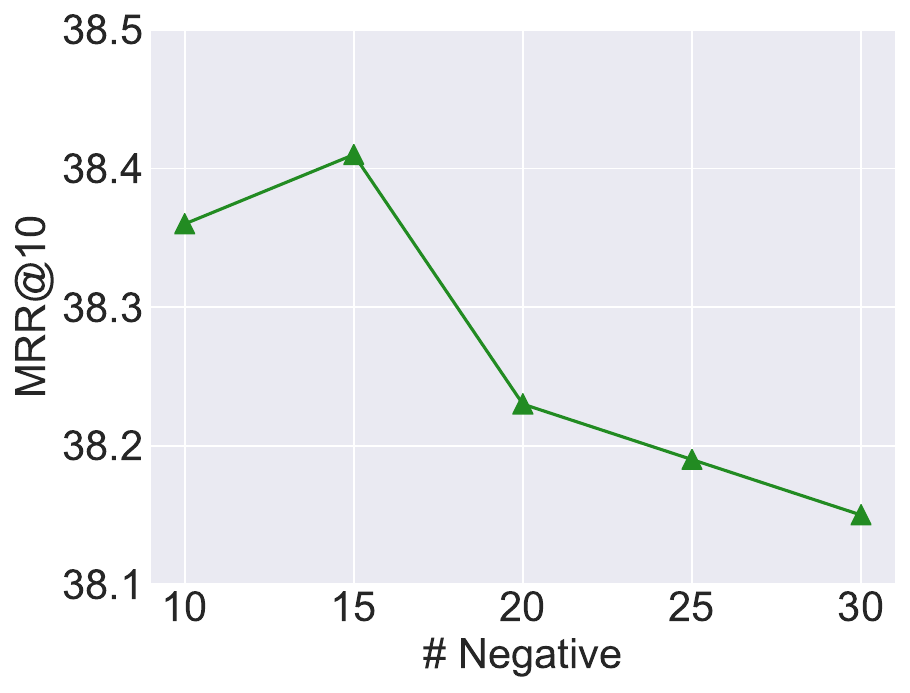}
  } \hspace{-3mm}
  \subfigure[{\small Dist. of Ranker output}] { \label{fig:dist_new}
    \includegraphics[width=0.473\columnwidth]{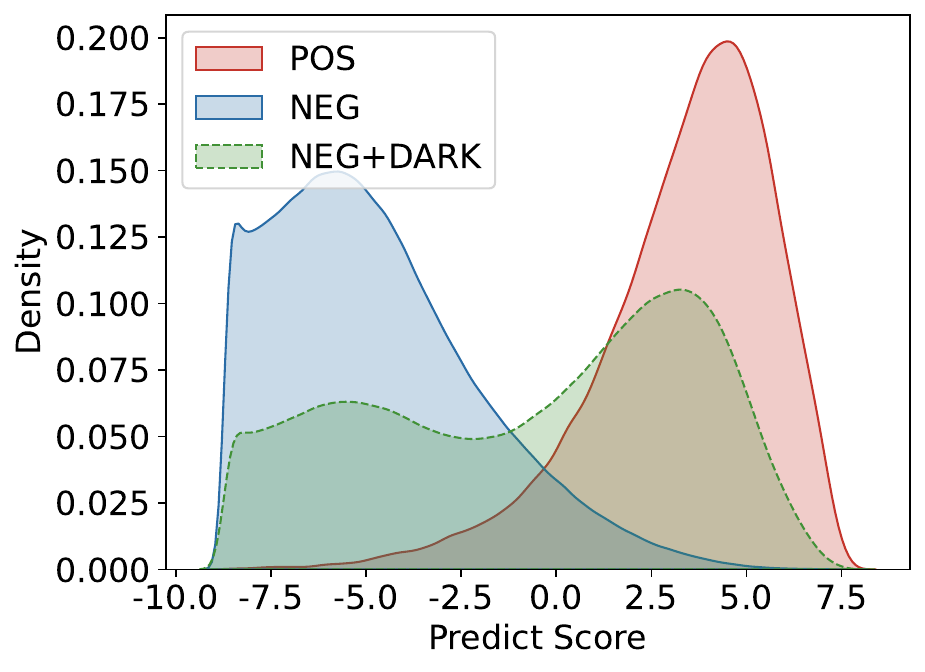}
  }
  \caption{(a) The impact of $m$; (b) Distributions of model prediction for the R$^2$anker over MS-MARCO.}
  \label{fig:turns-trend}
\end{figure}

\vspace{1mm}
{\textit{The impact of the number of negatives.}}
When constructing the training set, the number of negatives plays a vital role as it also indirectly controls the number of dark examples. 
To explore the effect of the number of negative samples as well as to find the best choice for $m$, we conduct experiments on different $m$\footnote{To better analyze the impact of the number of negative samples, we conduct the experiment on the model without adaptive dark examples.}. 
As illustrated in Figure~\ref{fig:trend_m}, when $m$ is small, increasing $m$ brings a positive effect and leads to the best performance when $m=15$. 
But as the curve indicates, incorporating more negatives brings no benefit, which is also in line with existing findings~\cite{Karpukhin2020DPR}. The above trend also indicates that too many trivial negatives ($m>15$) can not always bring improvement while incorporating our dark examples can still bring improvement to the knowledge transfer. The phenomenon also reveals the importance of distillation data in IR knowledge transfer.

\vspace{1mm}
\emph{The impact of dark examples on the output distribution of ranker.} Finally, we examine the impact of dark examples on the output distribution of the ranker. As illustrated in Figure \ref{fig:dist_new}, we draw the score distributions of the positive, negative candidates, and negative candidates plus dark examples using a teacher (R$^2$anker) over MS-MARCO. It can be observed that the scores for most original hard negatives are quite low and distributed far from the positives that have high scores. 
By incorporating these dark examples, we are able to improve the smoothness of the score distribution and prob our teacher model with a wider range of candidates that are more diversely relevant to the query. This enables us to more effectively transfer valuable "dark" knowledge from the teacher model.

\section{Conclusion}
\vspace{-1.5mm}
In this paper, we propose a knowledge distillation framework that can better transfer the dark knowledge in the cross-encoder with adaptive dark examples to help the dual-encoder achieve better performance.
We propose two approaches to create dark examples that are much harder for the cross-encoder teacher to distinguish than typical hard negatives to transfer more dark knowledge.
Further, we propose a self-paced distillation strategy that transfers the knowledge adaptively with high-confidence training instances. 
Experimental results in two widely-used benchmarks verify the effectiveness of our proposed method.

\section*{Limitations}
(i) \textit{Training computation overheads}: although having the same inference complexity as any other dense retrieval models, our approach requires more computation resources during training as it expands the number of negatives with the augmented dark examples. (ii) \textit{More analysis on noisy positives}: due to the limited computation resource, we only test and compare several typical settings of noisy positives, better strategies for constructing noisy positives (e.g., better masking methods and varying the number of noisy positives) can be explored to further improve the performance.

\bibliography{custom,anthology}

\begin{thebibliography}{69}
\expandafter\ifx\csname natexlab\endcsname\relax\def\natexlab#1{#1}\fi

\bibitem[{Cai et~al.(2022)Cai, Tao, Shen, Xu, Geng, Lin, He, and Jiang}]{cai2022hyper}
ZeFeng Cai, Chongyang Tao, Tao Shen, Can Xu, Xiubo Geng, Xin~Alex Lin, Liang He, and Daxin Jiang. 2022.
\newblock Hyper: Multitask hyper-prompted training enables large-scale retrieval generalization.
\newblock In \emph{The Eleventh International Conference on Learning Representations}.

\bibitem[{Chen et~al.(2017)Chen, Fisch, Weston, and Bordes}]{chen17openqa}
Danqi Chen, Adam Fisch, Jason Weston, and Antoine Bordes. 2017.
\newblock Reading wikipedia to answer open-domain questions.
\newblock In \emph{Proceedings of the 55th Annual Meeting of the Association for Computational Linguistics, {ACL} 2017, Vancouver, Canada, July 30 - August 4, Volume 1: Long Papers}, pages 1870--1879.

\bibitem[{Cohen et~al.(2024)Cohen, Indelman, Fairstein, and Kushilevitz}]{cohen2024indi}
Nachshon Cohen, Hedda~Cohen Indelman, Yaron Fairstein, and Guy Kushilevitz. 2024.
\newblock Indi: Informative and diverse sampling for dense retrieval.

\bibitem[{Craswell et~al.(2020)Craswell, Mitra, Yilmaz, Campos, and Voorhees}]{Craswell2020TREC19}
Nick Craswell, Bhaskar Mitra, Emine Yilmaz, Daniel Campos, and Ellen~M. Voorhees. 2020.
\newblock \href {http://arxiv.org/abs/2003.07820} {Overview of the {TREC} 2019 deep learning track}.
\newblock \emph{CoRR}, abs/2003.07820.

\bibitem[{Dai and Callan(2019{\natexlab{a}})}]{Dai2019DeepCT}
Zhuyun Dai and Jamie Callan. 2019{\natexlab{a}}.
\newblock \href {http://arxiv.org/abs/1910.10687} {Context-aware sentence/passage term importance estimation for first stage retrieval}.
\newblock \emph{CoRR}, abs/1910.10687.

\bibitem[{Dai and Callan(2019{\natexlab{b}})}]{deepct2019sigir}
Zhuyun Dai and Jamie Callan. 2019{\natexlab{b}}.
\newblock Deeper text understanding for {IR} with contextual neural language modeling.
\newblock In \emph{Proceedings of the 42nd International {ACM} {SIGIR} Conference on Research and Development in Information Retrieval, {SIGIR} 2019, Paris, France, July 21-25, 2019}, pages 985--988.

\bibitem[{Devlin et~al.(2019{\natexlab{a}})Devlin, Chang, Lee, and Toutanova}]{bert2019naacl}
Jacob Devlin, Ming{-}Wei Chang, Kenton Lee, and Kristina Toutanova. 2019{\natexlab{a}}.
\newblock {BERT:} pre-training of deep bidirectional transformers for language understanding.
\newblock In \emph{Proceedings of the 2019 Conference of the North American Chapter of the Association for Computational Linguistics: Human Language Technologies, {NAACL-HLT} 2019, Minneapolis, MN, USA, June 2-7, 2019, Volume 1 (Long and Short Papers)}, pages 4171--4186.

\bibitem[{Devlin et~al.(2019{\natexlab{b}})Devlin, Chang, Lee, and Toutanova}]{Devlin2019BERT}
Jacob Devlin, Ming{-}Wei Chang, Kenton Lee, and Kristina Toutanova. 2019{\natexlab{b}}.
\newblock \href {https://doi.org/10.18653/v1/n19-1423} {{BERT:} pre-training of deep bidirectional transformers for language understanding}.
\newblock In \emph{Proceedings of the 2019 Conference of the North American Chapter of the Association for Computational Linguistics: Human Language Technologies, {NAACL-HLT} 2019, Minneapolis, MN, USA, June 2-7, 2019, Volume 1 (Long and Short Papers)}, pages 4171--4186. Association for Computational Linguistics.

\bibitem[{Dinan et~al.(2018)Dinan, Roller, Shuster, Fan, Auli, and Weston}]{dinan2018wizard}
Emily Dinan, Stephen Roller, Kurt Shuster, Angela Fan, Michael Auli, and Jason Weston. 2018.
\newblock Wizard of wikipedia: Knowledge-powered conversational agents.
\newblock In \emph{International Conference on Learning Representations}.

\bibitem[{Formal et~al.(2021)Formal, Lassance, Piwowarski, and Clinchant}]{Formal2021SPLADEv2}
Thibault Formal, Carlos Lassance, Benjamin Piwowarski, and St{\'{e}}phane Clinchant. 2021.
\newblock \href {http://arxiv.org/abs/2109.10086} {{SPLADE} v2: Sparse lexical and expansion model for information retrieval}.
\newblock \emph{CoRR}, abs/2109.10086.

\bibitem[{Gao and Callan(2021{\natexlab{a}})}]{Gao2021Condenser}
Luyu Gao and Jamie Callan. 2021{\natexlab{a}}.
\newblock \href {https://doi.org/10.18653/v1/2021.emnlp-main.75} {Condenser: a pre-training architecture for dense retrieval}.
\newblock In \emph{Proceedings of the 2021 Conference on Empirical Methods in Natural Language Processing, {EMNLP} 2021, Virtual Event / Punta Cana, Dominican Republic, 7-11 November, 2021}, pages 981--993. Association for Computational Linguistics.

\bibitem[{Gao and Callan(2021{\natexlab{b}})}]{Gao2021coCondenser}
Luyu Gao and Jamie Callan. 2021{\natexlab{b}}.
\newblock \href {http://arxiv.org/abs/2108.05540} {Unsupervised corpus aware language model pre-training for dense passage retrieval}.
\newblock \emph{CoRR}, abs/2108.05540.

\bibitem[{Gao and Callan(2022)}]{gao2022unsupervised}
Luyu Gao and Jamie Callan. 2022.
\newblock Unsupervised corpus aware language model pre-training for dense passage retrieval.
\newblock In \emph{ACL}.

\bibitem[{Gao et~al.(2021)Gao, Dai, and Callan}]{gao2021coil}
Luyu Gao, Zhuyun Dai, and Jamie Callan. 2021.
\newblock {COIL}: Revisit exact lexical match in information retrieval with contextualized inverted list.
\newblock In \emph{Proceedings of the 2021 Conference of the North American Chapter of the Association for Computational Linguistics: Human Language Technologies}, pages 3030--3042.

\bibitem[{Gillick et~al.(2018)Gillick, Presta, and Tomar}]{gillick2018end}
Daniel Gillick, Alessandro Presta, and Gaurav~Singh Tomar. 2018.
\newblock End-to-end retrieval in continuous space.
\newblock \emph{arXiv preprint arXiv:1811.08008}.

\bibitem[{Guo et~al.(2019)Guo, Mao, and Zhang}]{guo2019augmenting}
Hongyu Guo, Yongyi Mao, and Richong Zhang. 2019.
\newblock Augmenting data with mixup for sentence classification: An empirical study.
\newblock \emph{arXiv preprint arXiv:1905.08941}.

\bibitem[{Henderson et~al.(2017)Henderson, Al{-}Rfou, Strope, Sung, Luk{\'{a}}cs, Guo, Kumar, Miklos, and Kurzweil}]{henderson2017efficient}
Matthew~L. Henderson, Rami Al{-}Rfou, Brian Strope, Yun{-}Hsuan Sung, L{\'{a}}szl{\'{o}} Luk{\'{a}}cs, Ruiqi Guo, Sanjiv Kumar, Balint Miklos, and Ray Kurzweil. 2017.
\newblock Efficient natural language response suggestion for smart reply.
\newblock \emph{CoRR}, abs/1705.00652.

\bibitem[{Hinton et~al.(2015)Hinton, Vinyals, and Dean}]{Hinton2015distilling}
Geoffrey~E. Hinton, Oriol Vinyals, and Jeffrey Dean. 2015.
\newblock Distilling the knowledge in a neural network.
\newblock \emph{CoRR}, abs/1503.02531.

\bibitem[{Hofst{\"{a}}tter et~al.(2020)Hofst{\"{a}}tter, Althammer, Schr{\"{o}}der, Sertkan, and Hanbury}]{hostatter2020improving}
Sebastian Hofst{\"{a}}tter, Sophia Althammer, Michael Schr{\"{o}}der, Mete Sertkan, and Allan Hanbury. 2020.
\newblock Improving efficient neural ranking models with cross-architecture knowledge distillation.
\newblock \emph{CoRR}, abs/2010.02666.

\bibitem[{Hofst{\"{a}}tter et~al.(2021{\natexlab{a}})Hofst{\"{a}}tter, Lin, Yang, Lin, and Hanbury}]{tas2021}
Sebastian Hofst{\"{a}}tter, Sheng{-}Chieh Lin, Jheng{-}Hong Yang, Jimmy Lin, and Allan Hanbury. 2021{\natexlab{a}}.
\newblock Efficiently teaching an effective dense retriever with balanced topic aware sampling.
\newblock \emph{CoRR}, abs/2104.06967.

\bibitem[{Hofst{\"{a}}tter et~al.(2021{\natexlab{b}})Hofst{\"{a}}tter, Lin, Yang, Lin, and Hanbury}]{Hofstatter2021TAS-B}
Sebastian Hofst{\"{a}}tter, Sheng{-}Chieh Lin, Jheng{-}Hong Yang, Jimmy Lin, and Allan Hanbury. 2021{\natexlab{b}}.
\newblock \href {https://doi.org/10.1145/3404835.3462891} {Efficiently teaching an effective dense retriever with balanced topic aware sampling}.
\newblock In \emph{{SIGIR} '21: The 44th International {ACM} {SIGIR} Conference on Research and Development in Information Retrieval, Virtual Event, Canada, July 11-15, 2021}, pages 113--122. {ACM}.

\bibitem[{Huang et~al.(2020)Huang, Sharma, Sun, Xia, Zhang, Pronin, Padmanabhan, Ottaviano, and Yang}]{Huang2020Embedding}
Jui{-}Ting Huang, Ashish Sharma, Shuying Sun, Li~Xia, David Zhang, Philip Pronin, Janani Padmanabhan, Giuseppe Ottaviano, and Linjun Yang. 2020.
\newblock Embedding-based retrieval in facebook search.
\newblock In \emph{{KDD} '20: The 26th {ACM} {SIGKDD} Conference on Knowledge Discovery and Data Mining, Virtual Event, CA, USA, August 23-27, 2020}, pages 2553--2561.

\bibitem[{Jang et~al.(2021)Jang, Kang, Hong, Myaeng, Park, Yoon, and Seo}]{uhdbert2021}
Kyoungrok Jang, Junmo Kang, Giwon Hong, Sung{-}Hyon Myaeng, Joohee Park, Taewon Yoon, and Hee{-}Cheol Seo. 2021.
\newblock {UHD-BERT:} bucketed ultra-high dimensional sparse representations for full ranking.
\newblock \emph{CoRR}, abs/2104.07198.

\bibitem[{Johnson et~al.(2019)Johnson, Douze, and J{\'e}gou}]{johnson2019billion}
Jeff Johnson, Matthijs Douze, and Herv{\'e} J{\'e}gou. 2019.
\newblock Billion-scale similarity search with gpus.
\newblock \emph{IEEE Transactions on Big Data}, 7(3):535--547.

\bibitem[{Kalantidis et~al.(2020)Kalantidis, Sariyildiz, Pion, Weinzaepfel, and Larlus}]{kalantidis2020hard}
Yannis Kalantidis, Mert~Bulent Sariyildiz, Noe Pion, Philippe Weinzaepfel, and Diane Larlus. 2020.
\newblock Hard negative mixing for contrastive learning.
\newblock \emph{Advances in Neural Information Processing Systems}, 33:21798--21809.

\bibitem[{Karpukhin et~al.(2020)Karpukhin, Oguz, Min, Lewis, Wu, Edunov, Chen, and Yih}]{Karpukhin2020DPR}
Vladimir Karpukhin, Barlas Oguz, Sewon Min, Patrick S.~H. Lewis, Ledell Wu, Sergey Edunov, Danqi Chen, and Wen{-}tau Yih. 2020.
\newblock \href {https://doi.org/10.18653/v1/2020.emnlp-main.550} {Dense passage retrieval for open-domain question answering}.
\newblock In \emph{Proceedings of the 2020 Conference on Empirical Methods in Natural Language Processing, {EMNLP} 2020, Online, November 16-20, 2020}, pages 6769--6781. Association for Computational Linguistics.

\bibitem[{Khattab and Zaharia(2020{\natexlab{a}})}]{colbert2020sigir}
Omar Khattab and Matei Zaharia. 2020{\natexlab{a}}.
\newblock Colbert: Efficient and effective passage search via contextualized late interaction over {BERT}.
\newblock In \emph{Proceedings of the 43rd International {ACM} {SIGIR} conference on research and development in Information Retrieval, {SIGIR} 2020, Virtual Event, China, July 25-30, 2020}, pages 39--48.

\bibitem[{Khattab and Zaharia(2020{\natexlab{b}})}]{Khattab2020ColBERT}
Omar Khattab and Matei Zaharia. 2020{\natexlab{b}}.
\newblock \href {https://doi.org/10.1145/3397271.3401075} {Colbert: Efficient and effective passage search via contextualized late interaction over {BERT}}.
\newblock In \emph{Proceedings of the 43rd International {ACM} {SIGIR} conference on research and development in Information Retrieval, {SIGIR} 2020, Virtual Event, China, July 25-30, 2020}, pages 39--48. {ACM}.

\bibitem[{Lee et~al.(2019)Lee, Chang, and Toutanova}]{Lee2019ICT}
Kenton Lee, Ming{-}Wei Chang, and Kristina Toutanova. 2019.
\newblock \href {https://doi.org/10.18653/v1/p19-1612} {Latent retrieval for weakly supervised open domain question answering}.
\newblock In \emph{Proceedings of the 57th Conference of the Association for Computational Linguistics, {ACL} 2019, Florence, Italy, July 28- August 2, 2019, Volume 1: Long Papers}, pages 6086--6096. Association for Computational Linguistics.

\bibitem[{Lin and Ma(2021)}]{Lin2021UniCOIL}
Jimmy Lin and Xueguang Ma. 2021.
\newblock \href {http://arxiv.org/abs/2106.14807} {A few brief notes on deepimpact, coil, and a conceptual framework for information retrieval techniques}.
\newblock \emph{CoRR}, abs/2106.14807.

\bibitem[{Lin et~al.(2023)Lin, Gong, Liu, Zhang, Lin, Dong, Jiao, Lu, Jiang, Majumder et~al.}]{lin2023prod}
Zhenghao Lin, Yeyun Gong, Xiao Liu, Hang Zhang, Chen Lin, Anlei Dong, Jian Jiao, Jingwen Lu, Daxin Jiang, Rangan Majumder, et~al. 2023.
\newblock Prod: Progressive distillation for dense retrieval.
\newblock In \emph{Proceedings of the ACM Web Conference 2023}, pages 3299--3308.

\bibitem[{Liu et~al.(2019)Liu, Ott, Goyal, Du, Joshi, Chen, Levy, Lewis, Zettlemoyer, and Stoyanov}]{Liu2019RoBERTa}
Yinhan Liu, Myle Ott, Naman Goyal, Jingfei Du, Mandar Joshi, Danqi Chen, Omer Levy, Mike Lewis, Luke Zettlemoyer, and Veselin Stoyanov. 2019.
\newblock \href {http://arxiv.org/abs/1907.11692} {Roberta: {A} robustly optimized {BERT} pretraining approach}.
\newblock \emph{CoRR}, abs/1907.11692.

\bibitem[{Loshchilov and Hutter(2017)}]{Loshchilov2017FixingWD}
Ilya Loshchilov and Frank Hutter. 2017.
\newblock Fixing weight decay regularization in adam.
\newblock \emph{ArXiv}, abs/1711.05101.

\bibitem[{Lu et~al.(2020)Lu, {\'{A}}brego, Ma, Ni, and Yang}]{negative2020google}
Jing Lu, Gustavo~Hern{\'{a}}ndez {\'{A}}brego, Ji~Ma, Jianmo Ni, and Yinfei Yang. 2020.
\newblock Neural passage retrieval with improved negative contrast.
\newblock \emph{CoRR}, abs/2010.12523.

\bibitem[{Lu et~al.(2021)Lu, Xiong, He, Ke, Malik, Dou, Bennett, Liu, and Overwijk}]{Lu2021SeedEncoder}
Shuqi Lu, Chenyan Xiong, Di~He, Guolin Ke, Waleed Malik, Zhicheng Dou, Paul Bennett, Tie{-}Yan Liu, and Arnold Overwijk. 2021.
\newblock \href {http://arxiv.org/abs/2102.09206} {Less is more: Pre-training a strong siamese encoder using a weak decoder}.
\newblock \emph{CoRR}, abs/2102.09206.

\bibitem[{Lu et~al.(2022)Lu, Liu, Liu, Shi, Huang, Sun, Tian, Wu, Wang, Yin et~al.}]{lu2022ernie}
Yuxiang Lu, Yiding Liu, Jiaxiang Liu, Yunsheng Shi, Zhengjie Huang, Shikun Feng~Yu Sun, Hao Tian, Hua Wu, Shuaiqiang Wang, Dawei Yin, et~al. 2022.
\newblock Ernie-search: Bridging cross-encoder with dual-encoder via self on-the-fly distillation for dense passage retrieval.
\newblock \emph{arXiv preprint arXiv:2205.09153}.

\bibitem[{Luan et~al.(2021{\natexlab{a}})Luan, Eisenstein, Toutanova, and Collins}]{Luan2021MEBERT}
Yi~Luan, Jacob Eisenstein, Kristina Toutanova, and Michael Collins. 2021{\natexlab{a}}.
\newblock \href {https://doi.org/10.1162/tacl\_a\_00369} {Sparse, dense, and attentional representations for text retrieval}.
\newblock \emph{Trans. Assoc. Comput. Linguistics}, 9:329--345.

\bibitem[{Luan et~al.(2021{\natexlab{b}})Luan, Eisenstein, Toutanova, and Collins}]{MEBERT}
Yi~Luan, Jacob Eisenstein, Kristina Toutanova, and Michael Collins. 2021{\natexlab{b}}.
\newblock Sparse, dense, and attentional representations for text retrieval.
\newblock \emph{Transactions of the Association for Computational Linguistics}, 9:329--345.

\bibitem[{Menon et~al.(2022)Menon, Jayasumana, Rawat, Kim, Reddi, and Kumar}]{menon2022defense}
Aditya Menon, Sadeep Jayasumana, Ankit~Singh Rawat, Seungyeon Kim, Sashank Reddi, and Sanjiv Kumar. 2022.
\newblock In defense of dual-encoders for neural ranking.
\newblock In \emph{International Conference on Machine Learning}, pages 15376--15400. PMLR.

\bibitem[{Nguyen et~al.(2016)Nguyen, Rosenberg, Song, Gao, Tiwary, Majumder, and Deng}]{Nguyen2016MSMARCO}
Tri Nguyen, Mir Rosenberg, Xia Song, Jianfeng Gao, Saurabh Tiwary, Rangan Majumder, and Li~Deng. 2016.
\newblock \href {http://ceur-ws.org/Vol-1773/CoCoNIPS\_2016\_paper9.pdf} {{MS} {MARCO:} {A} human generated machine reading comprehension dataset}.
\newblock In \emph{Proceedings of the Workshop on Cognitive Computation: Integrating neural and symbolic approaches 2016 co-located with the 30th Annual Conference on Neural Information Processing Systems {(NIPS} 2016), Barcelona, Spain, December 9, 2016}, volume 1773 of \emph{{CEUR} Workshop Proceedings}. CEUR-WS.org.

\bibitem[{Nogueira et~al.(2019{\natexlab{a}})Nogueira, Lin, and Epistemic}]{doctttttquery}
Rodrigo Nogueira, Jimmy Lin, and AI~Epistemic. 2019{\natexlab{a}}.
\newblock From doc2query to doctttttquery.
\newblock \emph{Online preprint}.

\bibitem[{Nogueira et~al.(2019{\natexlab{b}})Nogueira, Yang, Lin, and Cho}]{doc2query}
Rodrigo Nogueira, Wei Yang, Jimmy Lin, and Kyunghyun Cho. 2019{\natexlab{b}}.
\newblock Document expansion by query prediction.
\newblock \emph{CoRR}, abs/1904.08375.

\bibitem[{Park et~al.(2019)Park, Kim, Lu, and Cho}]{park2019relational}
Wonpyo Park, Dongju Kim, Yan Lu, and Minsu Cho. 2019.
\newblock Relational knowledge distillation.
\newblock In \emph{Proceedings of the IEEE/CVF Conference on Computer Vision and Pattern Recognition}, pages 3967--3976.

\bibitem[{Qu et~al.(2021)Qu, Ding, Liu, Liu, Ren, Zhao, Dong, Wu, and Wang}]{Qu2021RocketQA}
Yingqi Qu, Yuchen Ding, Jing Liu, Kai Liu, Ruiyang Ren, Wayne~Xin Zhao, Daxiang Dong, Hua Wu, and Haifeng Wang. 2021.
\newblock \href {https://doi.org/10.18653/v1/2021.naacl-main.466} {Rocketqa: An optimized training approach to dense passage retrieval for open-domain question answering}.
\newblock In \emph{Proceedings of the 2021 Conference of the North American Chapter of the Association for Computational Linguistics: Human Language Technologies, {NAACL-HLT} 2021, Online, June 6-11, 2021}, pages 5835--5847. Association for Computational Linguistics.

\bibitem[{Ren et~al.(2021{\natexlab{a}})Ren, Lv, Qu, Liu, Zhao, She, Wu, Wang, and Wen}]{pair}
Ruiyang Ren, Shangwen Lv, Yingqi Qu, Jing Liu, Wayne~Xin Zhao, QiaoQiao She, Hua Wu, Haifeng Wang, and Ji-Rong Wen. 2021{\natexlab{a}}.
\newblock {PAIR}: Leveraging passage-centric similarity relation for improving dense passage retrieval.
\newblock In \emph{Findings of the Association for Computational Linguistics: ACL-IJCNLP 2021}, pages 2173--2183.

\bibitem[{Ren et~al.(2021{\natexlab{b}})Ren, Lv, Qu, Liu, Zhao, She, Wu, Wang, and Wen}]{Ren2021PAIR}
Ruiyang Ren, Shangwen Lv, Yingqi Qu, Jing Liu, Wayne~Xin Zhao, Qiaoqiao She, Hua Wu, Haifeng Wang, and Ji{-}Rong Wen. 2021{\natexlab{b}}.
\newblock \href {https://doi.org/10.18653/v1/2021.findings-acl.191} {{PAIR:} leveraging passage-centric similarity relation for improving dense passage retrieval}.
\newblock In \emph{Findings of the Association for Computational Linguistics: {ACL/IJCNLP} 2021, Online Event, August 1-6, 2021}, volume {ACL/IJCNLP} 2021 of \emph{Findings of {ACL}}, pages 2173--2183. Association for Computational Linguistics.

\bibitem[{Ren et~al.(2021{\natexlab{c}})Ren, Qu, Liu, Zhao, She, Wu, Wang, and Wen}]{Ren2021RocketQAv2}
Ruiyang Ren, Yingqi Qu, Jing Liu, Wayne~Xin Zhao, Qiaoqiao She, Hua Wu, Haifeng Wang, and Ji{-}Rong Wen. 2021{\natexlab{c}}.
\newblock \href {https://doi.org/10.18653/v1/2021.emnlp-main.224} {Rocketqav2: {A} joint training method for dense passage retrieval and passage re-ranking}.
\newblock In \emph{Proceedings of the 2021 Conference on Empirical Methods in Natural Language Processing, {EMNLP} 2021, Virtual Event / Punta Cana, Dominican Republic, 7-11 November, 2021}, pages 2825--2835. Association for Computational Linguistics.

\bibitem[{Robertson and Zaragoza(2009)}]{Robertson2009BM25}
Stephen~E. Robertson and Hugo Zaragoza. 2009.
\newblock \href {https://doi.org/10.1561/1500000019} {The probabilistic relevance framework: {BM25} and beyond}.
\newblock \emph{Found. Trends Inf. Retr.}, 3(4):333--389.

\bibitem[{Romero et~al.(2014)Romero, Ballas, Kahou, Chassang, Gatta, and Bengio}]{romero2014fitnets}
Adriana Romero, Nicolas Ballas, Samira~Ebrahimi Kahou, Antoine Chassang, Carlo Gatta, and Yoshua Bengio. 2014.
\newblock Fitnets: Hints for thin deep nets.
\newblock \emph{arXiv preprint arXiv:1412.6550}.

\bibitem[{Santhanam et~al.(2021)Santhanam, Khattab, Saad{-}Falcon, Potts, and Zaharia}]{Khattab2021ColBERTv2}
Keshav Santhanam, Omar Khattab, Jon Saad{-}Falcon, Christopher Potts, and Matei Zaharia. 2021.
\newblock \href {http://arxiv.org/abs/2112.01488} {Colbertv2: Effective and efficient retrieval via lightweight late interaction}.
\newblock \emph{CoRR}, abs/2112.01488.

\bibitem[{Shen et~al.(2023)Shen, Geng, Tao, Xu, Long, Zhang, and Jiang}]{shen2023unifier}
Tao Shen, Xiubo Geng, Chongyang Tao, Can Xu, Guodong Long, Kai Zhang, and Daxin Jiang. 2023.
\newblock \href {https://doi.org/10.1145/3580305.3599927} {Unifier: A unified retriever for large-scale retrieval}.
\newblock In \emph{Proceedings of the 29th ACM SIGKDD Conference on Knowledge Discovery and Data Mining}, KDD '23, page 4787–4799, New York, NY, USA. Association for Computing Machinery.

\bibitem[{Thakur et~al.(2021)Thakur, Reimers, R{\"{u}}ckl{\'{e}}, Srivastava, and Gurevych}]{Thakur2021BEIR}
Nandan Thakur, Nils Reimers, Andreas R{\"{u}}ckl{\'{e}}, Abhishek Srivastava, and Iryna Gurevych. 2021.
\newblock \href {http://arxiv.org/abs/2104.08663} {{BEIR:} {A} heterogenous benchmark for zero-shot evaluation of information retrieval models}.
\newblock \emph{CoRR}, abs/2104.08663.

\bibitem[{Vaswani et~al.(2017)Vaswani, Shazeer, Parmar, Uszkoreit, Jones, Gomez, Kaiser, and Polosukhin}]{vaswani2017attention}
Ashish Vaswani, Noam Shazeer, Niki Parmar, Jakob Uszkoreit, Llion Jones, Aidan~N Gomez, {\L}ukasz Kaiser, and Illia Polosukhin. 2017.
\newblock Attention is all you need.
\newblock \emph{Advances in neural information processing systems}, 30.

\bibitem[{Wang et~al.(2023)Wang, Yang, Huang, Jiao, Yang, Jiang, Majumder, and Wei}]{wang2022simlm}
Liang Wang, Nan Yang, Xiaolong Huang, Binxing Jiao, Linjun Yang, Daxin Jiang, Rangan Majumder, and Furu Wei. 2023.
\newblock \href {https://doi.org/10.18653/v1/2023.acl-long.125} {{S}im{LM}: Pre-training with representation bottleneck for dense passage retrieval}.
\newblock In \emph{Proceedings of the 61st Annual Meeting of the Association for Computational Linguistics (Volume 1: Long Papers)}, pages 2244--2258, Toronto, Canada. Association for Computational Linguistics.

\bibitem[{Xiao et~al.(2022)Xiao, Liu, Shao, and Cao}]{xiao2022retromae}
Shitao Xiao, Zheng Liu, Yingxia Shao, and Zhao Cao. 2022.
\newblock \href {https://doi.org/10.18653/v1/2022.emnlp-main.35} {{R}etro{MAE}: Pre-training retrieval-oriented language models via masked auto-encoder}.
\newblock In \emph{Proceedings of the 2022 Conference on Empirical Methods in Natural Language Processing}, pages 538--548, Abu Dhabi, United Arab Emirates. Association for Computational Linguistics.

\bibitem[{Xiong et~al.(2020)Xiong, Xiong, Li, Tang, Liu, Bennett, Ahmed, and Overwijk}]{ANCE}
Lee Xiong, Chenyan Xiong, Ye~Li, Kwok{-}Fung Tang, Jialin Liu, Paul Bennett, Junaid Ahmed, and Arnold Overwijk. 2020.
\newblock Approximate nearest neighbor negative contrastive learning for dense text retrieval.
\newblock \emph{CoRR}, abs/2007.00808.

\bibitem[{Xiong et~al.(2021)Xiong, Xiong, Li, Tang, Liu, Bennett, Ahmed, and Overwijk}]{Xiong2021ANCE}
Lee Xiong, Chenyan Xiong, Ye~Li, Kwok{-}Fung Tang, Jialin Liu, Paul~N. Bennett, Junaid Ahmed, and Arnold Overwijk. 2021.
\newblock \href {https://openreview.net/forum?id=zeFrfgyZln} {Approximate nearest neighbor negative contrastive learning for dense text retrieval}.
\newblock In \emph{9th International Conference on Learning Representations, {ICLR} 2021, Virtual Event, Austria, May 3-7, 2021}. OpenReview.net.

\bibitem[{Xu et~al.(2018)Xu, Park, Yi, and Sutton}]{xu2018interpreting}
Kai Xu, Dae~Hoon Park, Chang Yi, and Charles Sutton. 2018.
\newblock Interpreting deep classifier by visual distillation of dark knowledge.
\newblock \emph{arXiv preprint arXiv:1803.04042}.

\bibitem[{Xu et~al.(2024)Xu, Li, Tao, Shen, Cheng, Li, Xu, Tao, and Zhou}]{xu2024survey}
Xiaohan Xu, Ming Li, Chongyang Tao, Tao Shen, Reynold Cheng, Jinyang Li, Can Xu, Dacheng Tao, and Tianyi Zhou. 2024.
\newblock A survey on knowledge distillation of large language models.
\newblock \emph{arXiv preprint arXiv:2402.13116}.

\bibitem[{Yang et~al.(2017{\natexlab{a}})Yang, Fang, and Lin}]{bm25}
Peilin Yang, Hui Fang, and Jimmy Lin. 2017{\natexlab{a}}.
\newblock Anserini: Enabling the use of lucene for information retrieval research.
\newblock In \emph{Proceedings of the 40th International {ACM} {SIGIR} Conference on Research and Development in Information Retrieval, Shinjuku, Tokyo, Japan, August 7-11, 2017}, pages 1253--1256.

\bibitem[{Yang et~al.(2017{\natexlab{b}})Yang, Fang, and Lin}]{Yang2017Anserini}
Peilin Yang, Hui Fang, and Jimmy Lin. 2017{\natexlab{b}}.
\newblock \href {https://doi.org/10.1145/3077136.3080721} {Anserini: Enabling the use of lucene for information retrieval research}.
\newblock In \emph{Proceedings of the 40th International {ACM} {SIGIR} Conference on Research and Development in Information Retrieval, Shinjuku, Tokyo, Japan, August 7-11, 2017}, pages 1253--1256. {ACM}.

\bibitem[{Yang et~al.(2020)Yang, Jin, Lin, Guo, and Cer}]{google2020augmentation}
Yinfei Yang, Ning Jin, Kuo Lin, Mandy Guo, and Daniel Cer. 2020.
\newblock Neural retrieval for question answering with cross-attention supervised data augmentation.
\newblock \emph{CoRR}, abs/2009.13815.

\bibitem[{Zeng et~al.(2022)Zeng, Zamani, and Vinay}]{zeng2022curriculum}
Hansi Zeng, Hamed Zamani, and Vishwa Vinay. 2022.
\newblock Curriculum learning for dense retrieval distillation.
\newblock In \emph{Proceedings of the 45th International ACM SIGIR Conference on Research and Development in Information Retrieval}, pages 1979--1983.

\bibitem[{Zhan et~al.(2021{\natexlab{a}})Zhan, Mao, Liu, Guo, Zhang, and Ma}]{Zhan2021STAR-ADORE}
Jingtao Zhan, Jiaxin Mao, Yiqun Liu, Jiafeng Guo, Min Zhang, and Shaoping Ma. 2021{\natexlab{a}}.
\newblock \href {https://doi.org/10.1145/3404835.3462880} {Optimizing dense retrieval model training with hard negatives}.
\newblock In \emph{{SIGIR} '21: The 44th International {ACM} {SIGIR} Conference on Research and Development in Information Retrieval, Virtual Event, Canada, July 11-15, 2021}, pages 1503--1512. {ACM}.

\bibitem[{Zhan et~al.(2021{\natexlab{b}})Zhan, Mao, Liu, Guo, Zhang, and Ma}]{Optimizing2021sigir}
Jingtao Zhan, Jiaxin Mao, Yiqun Liu, Jiafeng Guo, Min Zhang, and Shaoping Ma. 2021{\natexlab{b}}.
\newblock Optimizing dense retrieval model training with hard negatives.
\newblock \emph{CoRR}, abs/2104.08051.

\bibitem[{Zhang et~al.(2022)Zhang, Gong, Shen, Lv, Duan, and Chen}]{Zhang2021AR2}
Hang Zhang, Yeyun Gong, Yelong Shen, Jiancheng Lv, Nan Duan, and Weizhu Chen. 2022.
\newblock \href {https://openreview.net/forum?id=MR7XubKUFB} {Adversarial retriever-ranker for dense text retrieval}.
\newblock In \emph{International Conference on Learning Representations}.

\bibitem[{Zhang et~al.(2023)Zhang, Tao, Shen, Xu, Geng, Jiao, and Jiang}]{zhang2023led}
Kai Zhang, Chongyang Tao, Tao Shen, Can Xu, Xiubo Geng, Binxing Jiao, and Daxin Jiang. 2023.
\newblock Led: Lexicon-enlightened dense retriever for large-scale retrieval.
\newblock In \emph{Proceedings of the ACM Web Conference 2023}, pages 3203--3213.

\bibitem[{Zhou et~al.(2023)Zhou, Shen, Geng, Tao, Xu, Long, Jiao, and Jiang}]{zhou2022towards}
Yucheng Zhou, Tao Shen, Xiubo Geng, Chongyang Tao, Can Xu, Guodong Long, Binxing Jiao, and Daxin Jiang. 2023.
\newblock \href {https://doi.org/10.18653/v1/2023.findings-acl.332} {Towards robust ranker for text retrieval}.
\newblock In \emph{Findings of the Association for Computational Linguistics: ACL 2023}, pages 5387--5401, Toronto, Canada. Association for Computational Linguistics.

\bibitem[{Zou et~al.(2021)Zou, Zhang, Cai, Ma, Cheng, Wang, Shi, Cheng, and Yin}]{Zou2021Baidu}
Lixin Zou, Shengqiang Zhang, Hengyi Cai, Dehong Ma, Suqi Cheng, Shuaiqiang Wang, Daiting Shi, Zhicong Cheng, and Dawei Yin. 2021.
\newblock \href {https://doi.org/10.1145/3447548.3467147} {Pre-trained language model based ranking in baidu search}.
\newblock In \emph{{KDD} '21: The 27th {ACM} {SIGKDD} Conference on Knowledge Discovery and Data Mining, Virtual Event, Singapore, August 14-18, 2021}, pages 4014--4022. {ACM}.

\end{thebibliography}


\end{document}